# Worst-Case Upper Bound for (1, 2)-QSAT

Minghao Yin

Department of Computer, Northeast Normal University, Changchun, China, 130117

ymh@nenu.edu.cn

**Abstract.** The rigorous theoretical analysis of the algorithm for a subclass of QSAT, i.e. (1, 2)-QSAT, has been proposed in the literature. (1, 2)-QSAT, first introduced in SAT'08, can be seen as quantified extended 2-CNF formulas. Until now, within our knowledge, there exists no algorithm presenting the worst upper bound for (1, 2)-QSAT. Therefore in this paper, we present an exact algorithm to solve (1, 2)-QSAT. By analyzing the algorithms, we obtain a worst-case upper bound $O(1.4142^m)$, where m is the number of clauses.

## 1 Introduction

Quantified Boolean Formulas (QBF) can be viewed as a generalization version of the Boolean satisfiability (SAT) with existential or universal quantifiers to every variable in propositional formulas. This problem permits both universal and existential quantifiers over Boolean variables. Deciding whether a QBF is satisfiable is the prototypical problem for PSPACE-complete problems, and is also one of the most important issues in artificial intelligence field. Many practical problems can be transformed into QBF, such as intelligent planning problems, non-monotonic reasoning problems and reasoning about knowledge problems.

Based on the fact that there is no polynomial time algorithm exists for SAT problems, assuming P≠NP stands, developing fast exact algorithm for testing satisfiability of propositional formulas are crucial in determining the scale of the problems that can be solved. Because a slight improvement from $O(c^k)$ to $O((c-\varepsilon)^k)$ may significantly increase the size of the problem being tractable, it is not surprising to view tremendous efforts made by researchers to designing exact algorithms providing better upper bound of SAT problems. Take 3-SAT problem for instance. The currently best upper bound for deterministic algorithm is $O(1.439^n)$ achieved by Kutzkov and Scheder [6], and that for local search algorithm is $O(1.32216^n)$ achieved by Rolf [7].

When it comes to QBF problem, the best upper bound of QBF problem in 3CNF (conjunctive normal form) is $O(1.619^m)$ where m is the number of clauses, as can be seen in ([8]. However, this upper bound is achieved under the restriction that the number of clauses should be equal to the number of variables. Moreover, Williams pointed out that when the clause-to-variable ratio approaches 2, the bound approaches $2^n$. In this paper, we focus on a subclass of closed quantified Boolean formulas, (1, 2)-QSAT, which was first introduced in [2]. (1, 2)-QSAT-formulas are in the form of $\forall X \exists Y \varphi(X,Y)$, where $X$ has $n_1$ variables, Y has $n_2$ variables, and $\varphi(X,Y)$ is a conjunction of 3-clauses, each of which containing exactly one universal literal and two existential ones. This problem can be seen as an extension version of 2SAT. An extreme situation is that when $n_1$ is of logarithmic order compared to the number of variables n, this problem can be solved in polynomial time. But in general case, this problem is co-NP-complete [2]. Creignou have studied the phase transition phenomenon of (1, 2)-QSAT [2, 3]. However, up to now, within our knowledge, no literals have studied the worst upper bound of the problem. Therefore in this paper, we focus on deriving a non-trivial worst case upper bound for (1, 2)-QSAT. Specifically, we present an algorithm to slove (1, 2)-QSAT. By analyzing the algorithm, we obtain a worst-case upper bounds $O(1.4142^m)$, where m is the number of clauses.

## 2 Basic Concepts

In this section, we recall some necessary concepts in this paper. We describe some definitions used in this paper. A *variable x* can take the values *true* or *false*. A *literal* of a variable is either the unnegated literal *x*, having the same truth value as the variable, or the negated literal $\neg x$, having the opposite truth value as the variable. A *clause* is a disjunction of literals, referred to as a *k-clause* if the clause is a disjunction on *k* literals. A *k*-SAT formula *F* in Conjunction Normal Form (CNF) is a conjunction of clauses, each of which contains exactly *k* literals. A (1, 2)-QSAT formula F is in the form of

$$\forall X \exists Y \varphi(X, Y)$$

where *X* has $n_1$ variables, Y has $n_2$ variables, and $\varphi(X,Y)$ is a conjunction of 3-clauses, each of which containing exactly one universal literal and two existential ones. We define *m* as the number of clauses in *F*, and *n* as the number of variables *F* contains. Note that (1, 2)-QSAT can be viewed as an extension version of 2SAT, because if we can get a formula of 2SAT by deleting the universal literal in each clause and the superfluous arbitrary quantifier of (1, 2)-QSAT. A *truth assignment* for F is a map that assigns each variable a value. A (1, 2)-QSAT formula is *satisfiable* if for each assignment of the variables X, there exists an assignment to the variables Y such that $\varphi$ is true.

## 3 Estimating the Running Time

In this section, we explain how to compute the complexity of our algorithms. We will use a notion called branching tree. The branching tree [5] is a hierarchical tree structure consisting of a set of nodes, and each node is labeled with a proposition formula. Consider a node labeled with a formula $F$, then its sons are labeled with the sub formulae $F_1, F_2,…, F_n$, with $F_i (1 \leq i \leq n)$ being obtained by assigning a value to one of the variables in $F$. According to the above definition we can see that the process of constructing a branching tree is the same as the process of executing DPLL-style algorithms, therefore, we make use of the branching tree to estimate the time complexity.

In the branching tree, every node has a branching tuple [4]. Suppose a node is labeled with $F_0$ and its children nodes are labeled with $F_1, F_2, …, F_k$. Then the branching tuple of that node is $(r_1, r_2,…, r_k)$, where $r_i=m(F_0)-m(F_i)$ ($m(F_0)$ represents the number of clauses of $F_0$). The value of the branching tuple of a node, is called branching number ($\lambda(r_1, r_2,…, r_k)$), which is obtained as the positive root of the following equation.

$$\sum_{i=1}^{k} x^{-r_i} = 1. \tag{1}$$

The branching number of the branching tree can be defined as the maximum branching number(max $\lambda (r_1, r_2,…, r_k)$)of nodes among the branching tree. There is a close relationship between the branching number of a branching tree and the running time (T($m$)) of DPLL-style algorithms. Here, we assume that DPLL-style algorithms performing on each node take polynomial time. Then we obtain the following inequality.

$$T(m) \leq (\max \lambda (r_1, r_2,…, r_k))^m \times \text{poly}(F) = \max \sum_{i=1}^{k} T(m-r_i))^m \times \text{poly}(F). \tag{2}$$

Where $m$ is the number of clauses of the formula $F$, ploy($F$) is the polynomial time executing on the node $F$, and

$$(r_1, r_2,…, r_k) = \sum_{i=1}^{k} T(m-r_i). \tag{3}$$

In addition, let $m$ be the number of clauses, $m_i$ be the number of clauses in the sub-formula $F_i$ ($1 \leq i \leq k$) of the formula $F$. If a QBF problem recursively solved by the DPLL-style algorithms, the time required doesn't increase, for

$$\sum_{i=1}^{k} T(m_i) \leq T(m) \text{ where } m = \sum_{i=1}^{k} m_i. \tag{4}$$

## 4  Transformation Rules

This section discusses the transformation rules used in our algorithm for solving (1, 2)-QSAT. The transformation rules are applied before choosing a variable to branch. According to the complexity analysis described above, we just need to take into the difference value between the number of clauses of the input formula and the number of clauses of the formulae obtained from it by branching. The larger of the difference value, the smaller the upper bound obtained. In order to obtain a better upper bound, we should reduce as many clauses as possible. Therefore, we introduce some transformation rules such that the simplified formulae contain a fewer number of clauses and are equi-satisfiable with the original formula.

**Trivial falsity rule.** A (1, 2)-QSAT $F$ is *false*, if there is a clause $C$ only containing universal variables.

**Unit clause rule.** Let $F$ be a (1, 2)-QSAT formula, if there is a unit clause $C=\{l_i\}$ whose the only variable is existential, then the unit literal $l_i$ must be assigned *true*.

Note that the unit clause rule only can be applicable to the existential variables. Similarly, the unit propagation rule is also applicable to the existential variables. In fact, if there is a unit clause whose only variable is universal, then the (1, 2)-QSAT is *false* based on the trivial falsity rule. Therefore, the unit clause rule and the unit propagation rule are ruled out for universal variables. In the following, we will describe the unit propagation rule.

**Unit propagation rule.** Let $F$ be a (1, 2)-QSAT formula, if $F$ contains a unit literal $l_i$ and the literal is existential, then the unit propagation rule executes the following.

(1) Remove all clauses containing the literal $l_i$ from $F$;

(2) Delete all occurrences of the negation of the literal $l_i$ from the other clauses;

(3) Perform the process as far as possible.

**Monotone literal rule.** Given a (1, 2)-QSAT $F$ and a monotone literal $l_i$ in $F$, if $l_i$ is universal, then $l_i$ must be assigned *false*; if $l_i$ is existential, then $l_i$ must be assigned *true*.

Notice that the trivial falsity rule enables us to judge their truth value directly. And the other transformation rules can simplify the formulae into simpler ones. Now we discuss the equivalence between the simplified formula and the original one.

**Theorem 1.** Given a (1, 2)-QSAT $F$ and a simplified formula $F'$ obtained by applying the unit clause rule, the unit propagation rule, and the monotone literal rule, $F$ and $F'$ are equi-satisfiable.

*Proof.* In order to prove $F$ and $F'$ are equi-satisfiable, we need to prove the three rules make the simplified formula equi-satisfiable with $F$ respectively.

(1) $F$ and $F'$ are equi-satisfiable where $F'$ is obtained from $F$ by using the unit clause rule.

Suppose $F'$ is satisfiable. Consider a satisfying assignment $I$ for $F'$. It is obvious that the assignment $I \cup \{l_i\}$ satisfies $F$. On the other hand, in all satisfying assignments to $F$, the truth assignment to $l_i$ must be *true*. And $F' \setminus F = \varnothing$. Therefore, each assignment satisfying $F$ also satisfies $F'$.

(2) *F* and *F'* are equi-satisfiable where *F'* is obtained from *F* by using the unit propagation rule.

We know that *F* is satisfiable if and only if *F'* is satisfiable, where *F'* is obtained from *F* by using unit clause rule from (1). As a matter of fact, the unit propagation rule is performed based on the unit clause rule which is applied repeatedly as new unit clauses are generated. Therefore, *F* and *F'* are equi-satisfiable.

(3) *F* and *F'* are equi-satisfiable where *F'* is obtained from *F* by using the monotone literal rule.

We prove it in two cases. The one is that $l_i$ is universal; the other case is that $l_i$ is existential. Let's consider the first case.

$l_i$ is universal. Suppose *F'* is satisfiable. Consider a satisfying assignment *I* for *F'*. The assignment *I* satisfies the clauses that originally contain $l_i$ in *F*. Since $l_i$ assigned *false* has no influence to the truth value of the clauses $l_i$ in, $I \cup \{\neg l_i\}$ satisfies *F*. Moreover, it is clear that $I \cup \{l_i\}$ also satisfies *F*. Therefore, if *F'* is satisfiable, *F* is also satisfiable. On the contrary, since $F' \setminus F = \varnothing$, each assignment satisfying *F* also satisfies *F'*.

$l_i$ is existential. We assume that there is a satisfying assignment *I* for *F'*. It is obvious that the assignment $I \cup \{l_i\}$ satisfies *F*. On the other hand, $F' \setminus F = \varnothing$. Thus, each assignment satisfying *F* also satisfies *F'*.

Therefore, *F* and *F'* are equi-satisfiable, where *F'* is obtained by applying the unit clause rule, the unit propagation rule, and the monotone literal rule. □

# 5   Algorithm $S_m$ for Solving (1, 2)-QSAT

In this section, we present the algorithm $S_m$ for solving (1, 2)-QSAT and prove an upper bound $O(1.4142^m)$. The basic idea of the algorithm is to choose a variable in the outermost quantification set to branch, firstly transforming 3-CNF clauses into 2-CNF ones, and then solving the 2-CNF QSAT directly. The algorithm $S_m$ has one subroutine, Function $Reduce_m$, which is to reduce $F$ as far as possible by using the trivial falsity rule, the unit clause rule, the unit propagation rule, and the monotone literal rule. In the following, we describe the Function $Reduce_m$ in Figure 1. In the function, Exist_literals (resp. Univer_literals) denotes the set of the existential literals (resp. universal literals); $\{C: l_i \in C\}$ denotes the clauses contain the literal $l_i$.

---

**Function** $Reduce_m$ ($F$)

($R_1$): **If** there is a clause $C$ containing only universal literals, **then return** *false*.
($R_2$): **For** all $C = \{l_i\}$ **and** $l_i \in$ Exist_literals,
    $F := F / \{C: l_i \in C\}$,
    $F := F / \{C: \neg l_i \in C\} \wedge \{C: \neg l_i \in C\} / \neg l_i$,
  **Endfor**
($R_3$): **If** there is a monotone literal $l_i$ in $F$ **and** $l_i \in$ Exist_literals, **then**
    $F := F / \{C: l_i \in C\}$,
    $F := F / \{C: \neg l_i \in C\} \wedge \{C: \neg l_i \in C\} / \neg l_i$,
  **Endif**
  **If** there is a monotone literal $l_i$ in $F$ **and** $l_i \in$ Univer_literals, **then**
    $F := F / \{C: l_i \in C\} \wedge \{C: l_i \in C\} / l_i$,
  **Endif**
($R_4$): **If** $F$ changed in ($R_1$) - ($R_3$), **then** $Reduce_m$ ($F$).
    **Return** $F$.

---

**Fig. 1.** Framework of Function $Reduce_m$.

Next, let's consider the algorithm $S_m$. Given a (1, 2)-QSAT $F$, the basic strategy of Davis-Putnam-Logemann-Loveland (DPLL) is to choose a variable $v_i$ in the outermost quantification set. As we know, each clause in the (1, 2)-QSAT formulae consists of exactly one universal literal and two existential literals, and the universal literal is earlier quantified than the two existential literals. In other words, if there is a universal variable in a clause, the variable must be chosen as a branch variable when the algorithm deals with the clause. In the following, we propose the framework of the algorithm $S_m$ in Figure 2. At first, $F$ is simplified by the Function *Reduce$_m$* (line 1). If the simplified $F$ is empty, then $S_m$ stops and returns *true*, which means the input formula $F$ is satisfiable (line 2). If the simplified $F$ contains an empty clause, then $S_m$ stops and returns *false*, which means the input formula $F$ is unsatisfiable (line 3). If the algorithm doesn't stop, it chooses variables obey the quantification order until the (1, 2)-QSAT formula doesn't contain 3-CNF clauses (line 4). If the chosen variable is an existential variable, we just need to make sure one of the two branches is satisfiable. And if the chosen variable is a universal variable, we need to make sure that both branches lead to a satisfying assignment. When there is no 3-CNF clause, the algorithm will solve the formula by $QSAT_2(F)$. Note that the function $QSAT_2(F)$ can solve the 2-CNF QSAT in polynomial time [1].

---
*Algorithm* S$_m$(*F*)
---
*1.* *F*= *Reduce$_m$*(*F*).
*2*. **If** *F* is empty, **then return** *true*.
*3*. **If** *F* contains an empty clause, **then return** *false*.
*4.* **If** there exist 3-CNF clauses, pick a variable and construct two branches:
    (1) **If** there exists clauses of the form { $l_i$, $l_j$, $l_k$ } and { $\neg l_i$, $l'_j$, $l'_k$ },where $l_i$ is an universal literal,
        **then return** S$_\mathbf{m}$(*Reduce$_m$*(*F*[$l_i$])) $\wedge$ S$_\mathbf{m}$(*Reduce$_m$*(*F*[$\neg l_i$])).
    (2) **If** there exists clauses of the form { $l_i$, $l_j$, $l_k$ }, { $\neg l_i$, $l'_j$ }, and {$l$, $l'$, $\neg l'_j$ },where $l_i$ is an universal literal,
        **then return** S$_\mathbf{m}$(*Reduce$_m$*(*F*[$l_i$])) $\wedge$ S$_\mathbf{m}$(*Reduce$_m$*(*F*[$\neg l_i$])).
    (3) **If** there exists clauses of the form { $l_i$, $l_j$, $l_k$ }, { $\neg l_i$, $l'_j$ }, and {$l$, $\neg l'_j$ },where $l_i$ is an universal literal,
        **then return** S$_\mathbf{m}$(*Reduce$_m$*(*F*[$l_i$])) $\wedge$ S$_\mathbf{m}$(*Reduce$_m$*(*F*[$\neg l_i$])).
*5.* **Return** QSAT$_2$(*F*).
---

**Fig. 2.** Framework of Algorithm S$_m$.

# 6 Complexity Analysis

In this section, we use the branching tree to estimate the time complexity of the algorithm S$_m$. Since the function QSAT$_2$(*F*) can solve the 2-CNF QSAT in polynomial time, we only need to estimate the running time taken in the process of transformation from 3-clause formula into 2-clause formulae. The detailed proof will be presented in Theorem 2. Note that when computing the elements of the branching tuple, we only count the number of clauses without the universal variables by branching. This is because the simplified clauses containing only existential literals can be obtained by reducing or branching, but the simplified clauses containing universal literals can only be obtained by branching.

**Theorem 2.** Algorithm $S_m$ runs in $O(1.4142^m)$ time, where $m$ is the number of clauses.

*Proof.* Let us analyze the algorithm in detail.

Case 1 (line 1) is the course of simplification, which can solve in polynomial time. Thus it runs in $O(1)$.

Case 2 (line 2) and case 3 (line 3) can solve the problems completely. These cases run in $O(1)$.

In case 4.1, when $l_i=true$, every clause containing $l_i$ is removed and $\neg l_i$ is removed from clauses. That is to say at least the clause $\{l_i, l_j, l_k\}$ is removed and $\neg l_i$ is removed from the clause $\{\neg l_i, l'_j, l'_k\}$. Therefore, the current formula contains at least two 3-CNF clauses less than $F$. So we have $T(m)=2T(m-2)$ because the same situation is encountered when $l_i=false$. This case takes $O(1.4142^m)$ time.

In case 4.2, when $l_i=true$, $l'_j$ is forced to be assigned *true*. Thus, as least one clause is removed from $F$ and another 2-CNF clause is reduced to a clause consisting of only existential literal. When $l_i=false$, at lease two clauses are removed. Therefore, the worst case is when $T(m)=T(m-2)+T(m-2)$ with solution $O(1.4142^m)$.

In case 4.3, since $l_i=true$, $l'_j$ is forced to be assigned *true*. Then every clause containing $l_i$ or $l'_j$ is removed, and therefore, the current formula contains at least three clauses less than $F$. When $l_i=false$, every clause containing $\neg l_i$ is removed and $l_i$ is removed from clauses. Thus, at least one clause removed from $F$ and one 3-CNF clause is reduced to 2-CNF clause. Therefore, the worst case is when $T(m)=T(m-2)+T(m-2)$ with solution $O(1.4142^m)$.

In total, the upper bound for the algorithm $S_m$ is $O(1.4142^m)$. □

## 7  Conclusion

This paper addresses the worst-case upper bound for the (1, 2)-QSAT problem with the number of clauses as the parameter. The basic idea behind the algorithm is to transform 3-CNF clauses into 2-CNF ones, and then solving the 2-CNF QSAT directly. After a skillful analysis of these algorithms, we obtain the worst-case upper bound $O(1.4142^m)$ for X3SAT.

## Acknowledgments

This research is fully supported by the National Natural Science Foundation of China under Grant No.60803102, and also funded by NSFC Major Research Program 60496321: Basic Theory and Core Techniques of Non Canonical Knowledge.